\documentclass[runningheads]{llncs}
\usepackage{graphicx}
\usepackage{times}
\usepackage{latexsym}
\usepackage{bm}

\usepackage{cite}
\usepackage{amsmath,amssymb,amsfonts}

\usepackage{graphicx}
\usepackage{textcomp}
\usepackage{xcolor}

\usepackage{microtype}

\usepackage{latexsym}
\usepackage{enumerate}
\usepackage{multirow}
\usepackage{booktabs}
\usepackage{makecell}
\usepackage{wrapfig}
\usepackage{picins}

\usepackage{subfig}
\usepackage{amsmath}
\usepackage{algorithm}
\usepackage{algorithmic}
\usepackage{float}
\usepackage{framed} 
\usepackage[marginal]{footmisc}

\begin{document}
\title{HieNet: Bidirectional Hierarchy Framework for Automated ICD Coding}
%
%

\author{Shi Wang\inst{1*} \and
Daniel Tang\inst{2*} \and
Luchen Zhang\inst{3} \and
Huilin Li\inst{4} \and
Ding Han\inst{5}}

\authorrunning{F. Author et al.}
%
\institute{Key Laboratory of Intelligent Information Processing, Institute of Computing Technology,Chinese Academy of Sciences, China,
\email{wangshi@ict.ac.cn}\\ \and
University of Luxembourg, Interdisciplinary Centre for Security, Reliability and Trust (SNT), TruX, Luxembourg,
\email{xunzhu.tang@uni.lu}\\ \and
National Computer Network Emergency Response Technical Team/Coordination Center of China,
\email{zlc@cert.org.cn}\\ \and
Department of Civil Engineering, Technical University of Denmark, 2800 Lyngby, Denmark
\email{huillili57@gmail.com} \\ \and
Huazhong University of Science and Technology, China,
\email{dinghan93@gmail.com}}

\maketitle              
\footnote{* represents equal contribution to this work.\\}
\begin{abstract}
International Classification of Diseases (ICD) is a set of classification codes for medical records. Automated ICD coding, which assigns unique International Classification of Diseases codes with each medical record, is widely used recently for its efficiency and error-prone avoidance. However, there are challenges that remain such as heterogeneity, label unbalance, and complex relationships between ICD codes. In this work, we proposed a novel Bidirectional Hierarchy Framework(HieNet) to address the challenges. Specifically, a personalized PageRank routine is developed to capture the co-relation of codes, a bidirectional hierarchy passage encoder to capture the codes' hierarchical representations, and a progressive predicting method is then proposed to narrow down the semantic searching space of prediction. We validate our method on two widely used datasets. Experimental results on two authoritative public datasets demonstrate that our proposed method boosts the state-of-the-art performance by a large margin.

\end{abstract}

\section{Introduction}
The International Classification of Diseases (ICD) is widely considered as a healthcare multi-label classification system, supported by the World Health Organization (WHO). ICD codes have widely been used for reimbursement, taxonomy of diagnoses and procedures, and monitoring health issues \cite{nadathur2010maximising, DBLP:journals/midm/AvatiJHDNS18}. ICD coding needs coder to assign proper codes to a patient visit, which is composed of multiple long and heterogeneous textual narratives (e.g., discharge diagnosis, procedure notes, event notes), authored by different healthcare professionals, which means it's time-assuming, error-prone, and expensive in manual way. As a result, automated ICD encoding has attracted much attention since it can save time and labor for billing. A number of neural network methods handling automated coding as multi-label task, have been proposed by \cite{mullenbach-etal-2018-explainable, cao-etal-2020-hypercore, DBLP:conf/cikm/XieXYZ19}, which converts the ICD coding into a set of binary classification for each code. 

\begin{figure}[]
\centerline{\includegraphics[width=0.9\linewidth,keepaspectratio]{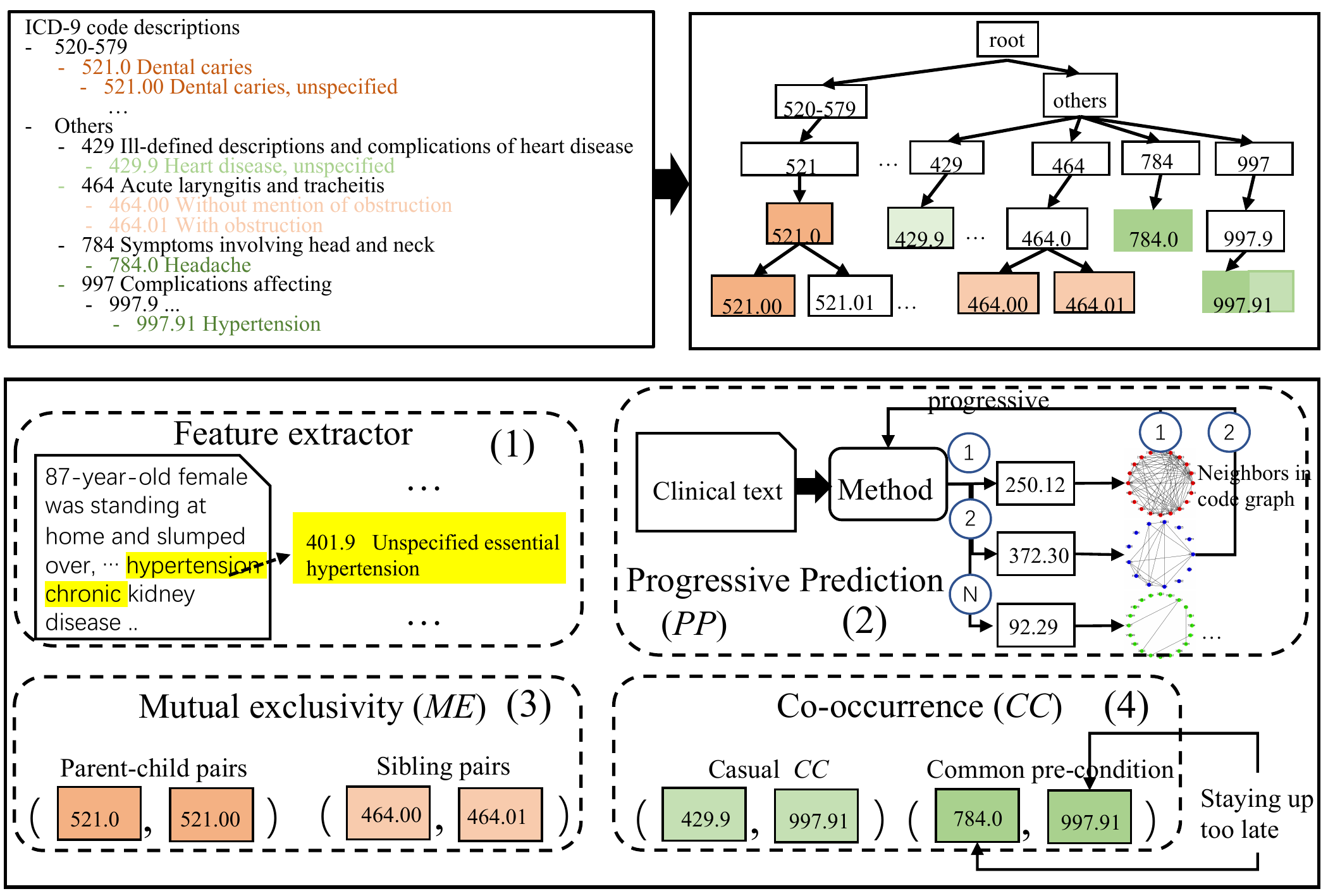}}
\caption{A schematic diagram of the ICD code tree, matching process between clinical notes and codes (1), progressive mechanism (2), and relationships among ICD codes ((3) and (4)).}
\label{fig: totree}
\end{figure}

ICD codes can be organized in a tree-like structure. If a node \emph{N} represents a kind of disease, the children of \emph{N} are the sub-types of this disease. And in many cases, the differences among the siblings from one parent disease are very subtle. There are several challenges to link the discharge summaries with ICD codes: (1) Each admission record has a long and complex discharge summary. And the summaries are usually more than one thousand words long, containing medical history, diagnosis texts, surgical procedures, etc. Thus, it is difficult to assign proper codes to a given clinical note. (2) The label space to predict of ICD codes is very large (e.g., over 18,000 for ICD-9-CM) and the label distribution is extremely unbalanced. 
However, the average length of training-label codes in MIMIC-\uppercase\expandafter{\romannumeral3}-full is only 15.89 and the length of codes in 80\% clinical notes is less than 22. Therefore, we only need to ensure the accuracy of top predicted codes. (3) Mutual exclusivity (\textit{ME}) in ICD codes:  As shown in Figure \ref{fig: totree}), we use deep and light orange colors to represent \textit{ME} codes. Deeper orange color represents `parent-child' codes, and lighter orange color represents sibling codes. For example, given a clinical text, if a finer code `521.00' is predicted, the parent of it `521.0' and grandparent `521' should not occur in the predicted results. Furthermore, some sibling ICD codes should not appear in the same predicted result, such as `464.00' (Acute laryngitis without mention of obstruction) and `464.01' (Acute laryngitis with obstruction), because they are anti-sense. (4) Co-occurrence (\textit{CC}) in ICD codes: We leverage light green and deep green colors for \textit{CC} codes. Light green color represents reasoning co-occurrence codes. For example, `997.91' (hypertension) usually leads to the occurrence of `429.9' (heart disease, unspecified); Deep green color represents \textit{CC} codes caused by common pre-conditions. For example, `staying up too late' usually causes `997.91' (hypertension) and `784.0' (headache).  

Some methods based on CNN (\cite{mullenbach-etal-2018-explainable, DBLP:conf/cikm/XieXYZ19, li2020icd}) were proposed to address the issues from the characteristic (1) above, and they were proved efficient in extracting features from long texts. For characteristic (2), to our knowledge, no previous scheme was proposed to solve it. To address problem with characteristic (3), \cite{xie2018neural} leverage a sequential tree-lstm architecture to extract structural features of code tree. However, as we know, there is no contextual relationships among siblings in the code tree and it's hard for tree-lstm process long sequential issue. Another approach \cite{cao-etal-2020-hypercore} leverages hyperbolic ball to encode the \textit{ME} feature, but it's hard to measure the real performance of this non-euclidean method on \textit{ME} problem. To address problem with characteristic (4), previous methods (e.g., \cite{cao-etal-2020-hypercore}) employ GCN \cite{DBLP:conf/iclr/KipfW17} to encode co-occurrence features of codes. However, GCN is unsuitable for describing the root node's neighborhood and not designed the causal co-occurrence cases. In summary, issues from characteristics (2), (3), (4) remain to be solved.

In this paper, to address issues above, we present a novel method HieNet, which is short for Bidirectional Hierarchy Framework for Automated ICD Coding. HieNet contains three main modules, including  a progressive mechanism (\textit{PM} module), a bidirectional hierarchy passage encoder (\textit{BHPE} module), and a personalized PageRank (\textit{PP} module). These modules are designed as the solution of (2), (3), and (4), respectively. 

\textbf{Progressive Prediction for (2)}: Given a clinical text or diagnose description, the previous methods predict all its ICD codes at the same time. However, some codes achieving high scores in their binary predictions can help predict other codes. For example, the neighbors of predicted code `401.9' (unspecified essential hypertension) in code tree usually contains some other potential gold labels, such as `348.4' (compression of brain). To make full use of first predicted, we introduce an approach named progressive mechanism to reduce the difficulty of improving the accuracy of predicted codes (e.g., average 14.3\% improvement on Jaccard metric over the average scores of the best baselines). 

\textbf{Hierarchy Features Encoding for (3)}: There are two patterns in mutual exclusivity (\textit{ME}): parent-child and sibling relationships. As stated above, ICD codes with \textit{ME} relationships should not occur in one clinical note. To address the issue, we propose a bidirectional hierarchy passage encoder (\textit{BHPE}) that contains two sub-modules: bidirectional passage retriever (\textit{BPR}) and tree position encoder (\textit{TPE}). The experimental results indicate that the \textit{BHPE} module improves HieNet by 12.0\% on macro-F1 on MIMIC-\uppercase\expandafter{\romannumeral3} full.

\textbf{Code Co-occurrence Encoding for (4)}: Some codes have causal or pre-condition co-occurrence relationship, which is called code co-occurrence (\textit{CC}). Pre-condition \textit{CC} codes are usually caused by common bad habits or hurts. We propose personalized PageRank (\textit{PP}) to encode pre-condition \textit{CC} features. Furthermore, the combination of \textit{PM} and \textit{PP} enable \textit{PP} has the ability to encode causal \textit{CC} relationships among ICD codes. The experimental results show that \textit{PP} module makes the  10.7\% improvement on macro-F1 on MIMIC-\uppercase\expandafter{\romannumeral3} full and 13.1\% improvement on macro-F1 on MIMIC-\uppercase\expandafter{\romannumeral2}.

\textbf{Our contributions}: 1) To the best of our knowledge, we are first to propose the progressive mechanism to improve the accuracy of $top_K$ predicted ICD codes. 2) We are first to introduce the bidirectional passage retriever and tree position encoder as the solutions of two patterns (i.e., parent-child, sibling) of mutual exclusivity. 3) We introduce a personalized PageRank to encode the pre-condition \textit{CC} and leverage progressive mechanism to capture casual \textit{CC} features. 4) The experimental results on two widely used datasets illustrate that HieNet outperforms the state-of-the-art compared to the previous methods (e.g., 20.6\% improvement on $top_{30}$-Jaccard (MIMIC-\uppercase\expandafter{\romannumeral3} 50) over the best baseline CAML).

\section{Related Work}

\textbf{Automatic ICD coding}: Automatic ICD coding has been studied in a large coverage of areas, including information retrieval, machine learning, and healthcare. \cite{kavuluru2015empirical} treat the ICD coding as a multi-label text classification problem and introduced a label ranking approach based on the features extracted from the clinical notes. \cite{mullenbach-etal-2018-explainable} proposed the landmark work CAML with attention algorithm and leveraged CNN to capture the key information for each code, and DR-CAML is a updated version of CAML with code description proposed in the same publication. Inspired by CAML, more CNN-based methods are proposed, including \cite{DBLP:conf/icml/AllamanisPS16,DBLP:journals/corr/abs-1710-00519,DBLP:journals/corr/SantosTXZ16}. \cite{shi2017towards} explored character based on LSTM with attention and \cite{xie2018neural} applied tree LSTM with hierarchy information for ICD coding. \cite{DBLP:conf/mlhc/XuLPGBMPKCMXX19} applied multi-modal machine learning to predict ICD codes. \cite{DBLP:conf/sigir/WangRCRN0R20} transferred the ICD coding into a path generation task and developed an adversarial reinforcement path generation framework for ICD coding. 

\textbf{Hierarchical Encoder}: Tree position encoder was proposed by \cite{DBLP:conf/aaai/FlorescuC17} and applied in tree-based transformers. Recent works demonstrate that tree position encoder can handle source code summarization \cite{DBLP:conf/acl/AhmadCRC20} and semantic prediction \cite{DBLP:conf/coling/HuberC20}. To our knowledge, we are the first to take the tree position encoder to capture the position embedding of code tree.

\textbf{Co-occurrence Encoder}: PageRank was proposed by \cite{page1999pagerank} and widely employed in website page ranking \cite{DBLP:journals/im/LangvilleM03}, latent topics mining \cite{DBLP:conf/acl/OguraK13}, key phrase extraction \cite{DBLP:conf/aaai/FlorescuC17}, and mutillingual word sense disambiguation \cite{DBLP:conf/acl/ScozzafavaMBTN20}. \cite{klicpera2018predict} proposed personalized PageRank (PPNP) to address the limit distribution of GCN. We are the first to use the personalized PageRank to encode the code co-occurrence representations for the automated ICD coding task. 

\section{Proposed Model}

\subsection{Problem Definition}

Following the previous works \cite{cao-etal-2020-hypercore, xie2018neural}, We formulate ICD coding as a multi-label text prediction problem. We make some definitions here to help state the proposed work well.

\begin{itemize}
    \item \textbf{Definition 1: \textit{CC}} \textit{CC} is short for Code co-occurrence and it contains three cases: (1) pre-condiction \textit{CC}. Intuitively, some diseases are usually caused by common bad habits; (2) causal \textit{CC}. (3) other co-occurrence codes without obvious reasons. 
    \item \textbf{Definition 2: \textit{ME}} \textit{ME} is short for Mutual exclusivity and it contains two cases: (1) Parent-child pairs. (2) Complementary sibling pairs, such as `464.00' (without mention of obstruction) and `464.01' (with obstruction).
    \item \textbf{Definition 3: \textit{DP}} \textit{DP} represents improving the accuracy of $top_K$ predicted codes. Actually, 80\% clinical notes have less than 22 codes, so we need to focus on improving the accuracy of $top_K$ predicted codes.
\end{itemize}

\subsection{Model Architecture}

\begin{figure*}[ht]
\centerline{\includegraphics[width=1\linewidth,keepaspectratio]{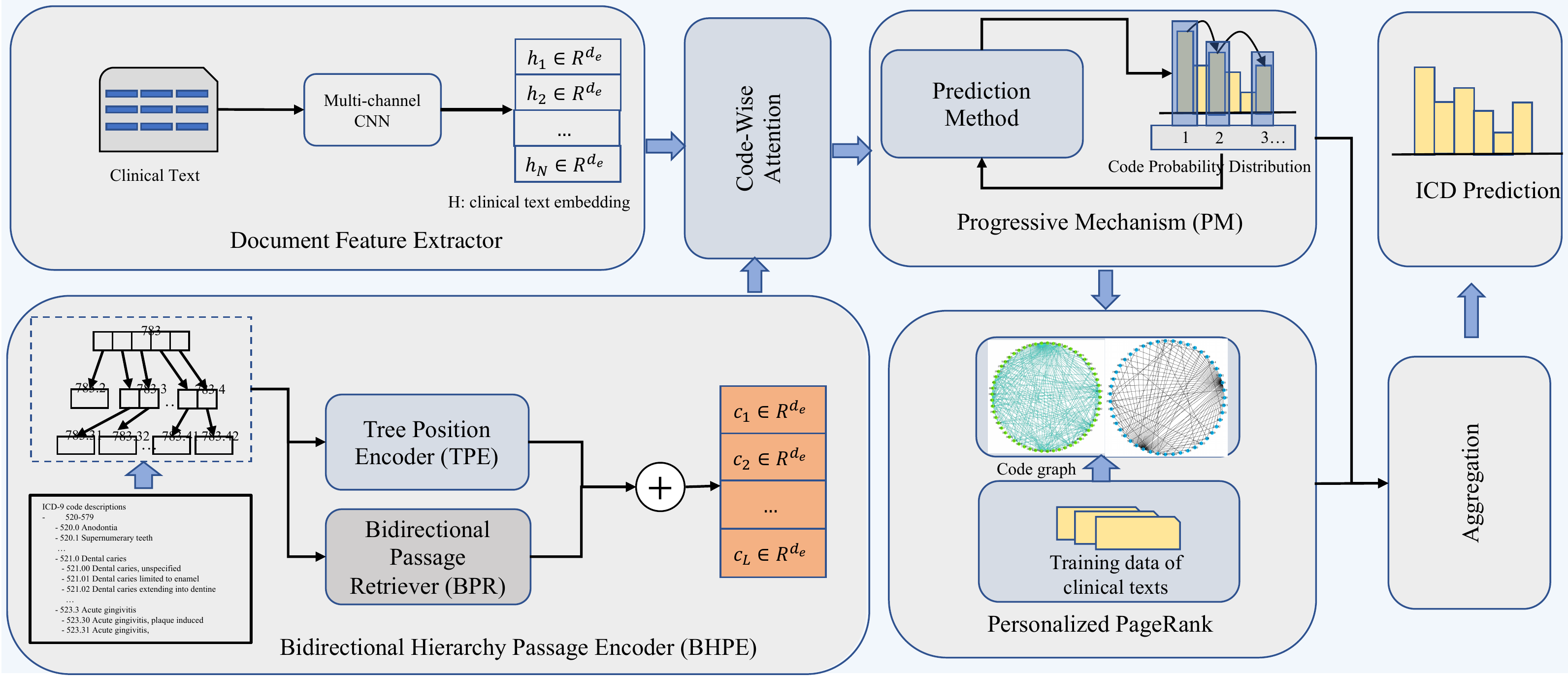}}
\caption{The architecture of our model.}
\label{fig: overview}
\end{figure*}

Figure \ref{fig: overview} shows an overview of the bidirectional hierarchy passage framework. Firstly, we encode the code hierarchy semantic as hierarchical code representations via bidirectional hierarchy passage retriever. Furthermore, we employ a multi-channel CNN to obtain clinical document embeddings and conduct a code-wise operation between document embeddings and hierarchical code representations. Secondly, we introduce a progressive mechanism to improve the accuracy of $top_K$ predicted ICD codes (\textbf{\textit{DP}}). Thirdly, we leverage a personalized PageRank algorithm to calculate the co-occurrence relationships among ICD codes. Finally, we aggregate the results of above modules and conduct a full connected layer with multiple sigmoid functions to generate 0-1 probability distributions for each code.

\subsection{Document Feature Extractor (\textit{DFE})} \label{sec: cnn}
Given a electronic health record {\it \textbf{W}}=$\lbrace$\bm{$w_1, w_2,...,w_N$} $\rbrace$ (\emph{N} donates the length of \textit{\textbf{W}}), we map \textit{\textbf{W}} into a vector representation \textit{\textbf{X}}$\in$ $\mathbb{R}^{d_e\times N}$ where $d_e$ indicates the dimension of word embedding. We leverage a multi-channel one-dimensional convolution neural network to encode clinical texts. A convolution operation contains a filter $\bm{W_f}$ with a window of \textit{k} words. For example, a feature $f_i$ is generated with a window of words $x_{i:i+k-1}$ by 
\small
\begin{equation}
    f_i = relu(W_f*x_{n:n+k-1}+b_c)
    \label{eq:cnnresult}
\end{equation}
where \textit{relu($\cdot$)} is a non-linear transformation function, \textit{b} $\in$ $\mathbb{R}^{d_e}$ is a bias, \textbf{*} is the concatenation operator. This filter is applied with multiple filter size to produce a final feature map \textit{f}:
\small
\begin{equation}
    f = f_{1,2,...,n-l+1} = f_1 \oplus f_2 \oplus \cdots \oplus f_{n-l+1}. \label{eq:filterresult}
\end{equation}

Then we employ a pooling operation over feature map \textit{f} and take the maximum value as the final value of \textit{f} by $\hat{f}$ = $\max(f)$. Regarding \textit{l} channels (\textit{l} different window sizes), we concatenate generated \textit{l} feature maps as a representation \textit{H} of an clinical text as follows:
\small
\begin{equation}
    H = \hat{f_1} \oplus \hat{f_2} \oplus \cdots \oplus \hat{f_{n-l+1}}. \label{eq:cnn}
\end{equation}

\subsection{Bidirectional Hierarchy Passage Encoder (\textit{BHPE})} \label{sec: BHE}

This section will introduce the hierarchy encoder that includes two main modules: bidirectional passage retriever (\textbf{\textit{BPR}}) and tree position encoder (\textbf{\textit{TPE}}). First, we construct a dimensional vector by averaging the vectors of words of code's description to represent the code. Then, we propose \textbf{\textit{BPR}} to capture parent-child relationships. Moreover, we introduce \textbf{\textit{TPE}} to encode tree positions of codes. 
Next, we add representations from \textit{BPR} and \textbf{\textit{TPE}} as the final vectors of codes 
that contain both hierarchical and parent-child contextual features. Finally, we obtain code-wise document representations by conducting code-wise attention between codes vectors and document representations.

\subsubsection{\textbf{Bidirectional Passage Retriever} (\textbf{\textit{BPR}})} \label{sec: BPR}
BPR uses two independent BERT encoders to capture the hierarchical relationships among ICD codes, including two directional process: up-stream \textit{u} (child $\rightarrow$ parent) and down-stream \textit{d} (parent $\rightarrow$ child):
\small
\begin{equation}
    \begin{split}
        e_u = BERT_u(u), e_d = BERT_d(d),
    \end{split}
\label{eq: udbert}
\end{equation}
where $e_u$ $\in$ $\mathbb{R}^{d}$ and $e_d$ $\in$ $\mathbb{R}^{d}$. We use the uncased version of BERT-base; therefore, \textit{d} = 768. The initial embedding tool of BERT we employ is WordPiece tokenization (\textit{wp}) which is different from the initial method of ICD codes (word2vec) \footnote{Why initial methods of ICD codes and clinical codes are diffident? Answer: ICD codes are usually beyond the vocabulary of BERT because they are professional and technical terms while words in clinical are original that can be covered by the vocabulary of BERT. Therefore, for better representing ICD codes and clinical notes, we leverage word2vec tool and WordPiece tokenization function in BERT to init them, respectively.}.

In the up-stream passage retriever, an internal node (with \textit{M} children) is comprised of these components: a position cell \textit{p} for each node, a self-input cell \textit{i$_\uparrow$}, and a BERT cell \{\textit{b$_\uparrow$}\}$_{m=1}^M$ for \textit{M} children. The position cell is used to encode the related relationship of ICD codes in the hierarchical code tree and the computation of \textit{p} is shown in section \ref{sec: spe}. The transition equations of among components are:
\small
\begin{equation}
    \begin{split}
        u = Set(p_k + wp(w_k))_{m=1}^M, \\
        i_\uparrow = {\{b_{\uparrow}\}}^M_{m=1} = BERT_u(u), 
    \end{split}
\label{eq: pib}
\end{equation}
where \textit{$w_k$} is the \textit{k-th} token of \textit{M} children of node \textit{C}. 

In the down-stream passage retriever, for a not-root node, it has such one component: an input cell \textit{i$_\downarrow$}. The transition equation is:
\small
\begin{equation}
    \begin{split}
        i_{\downarrow} = BERT_d(p + wp(w))
    \end{split}
\label{eq: dib}
\end{equation}
where \textit{p} is the position embedding of the parent node and \textit{w} is the average embedding of the node's description. Since root node has no parent, $i_\downarrow$ cannot be computed using the above equations. Instead, we set $i_\uparrow$ to $i_\downarrow$ here.

Loss function of \textbf{\textit{BPR}}: We call a pair of one up-stream \textit{PR} and one down-stream \textit{PR} an interaction. In one interaction, one node $n_j$ has two representations: $n_j^{i_\uparrow}$ and $n_j^{i_\downarrow}$. The goal of training \textbf{\textit{BPR}} is to reduce the difference between $n_j^{i_\uparrow}$ and $n_j^{i_\downarrow}$ as much as possible. Both $n_j^{i_\uparrow}$(\textit{X} = \{0, 1, 2,..., k,...,$d_e$\}) and $n_j^{i_\downarrow}$(X) can be recognized as two distributions. So we construct a Kullback-Leibler divergence \cite{peyre2015entropic} as the loss function of \textit{BPR}:
\small
\begin{equation}
    \begin{split}
        \mathcal{L}_{bpr1} &= \frac{1}{L}\Sigma_1^L\Sigma_1^{d_e}[n_j^{i_\uparrow}(x)log(n_j^{i_\uparrow}(x))-n_j^{i_\uparrow}(x)log(n_j^{i_\downarrow}(x))] \\
        \mathcal{L}_{bpr2} &= \frac{1}{L}\Sigma_1^L\Sigma_1^{d_e}[n_j^{i_\downarrow}(x)log(n_j^{i_\downarrow}(x))-n_j^{i_\downarrow}(x)log(n_j^{i_\uparrow}(x))] \\
        \mathcal{L}_{bpr} &= \mathcal{L}_{bpr1} + \mathcal{L}_{bpr2} + \lVert \mathcal{L}_{bpr1}-\mathcal{L}_{bpr2} \rVert^2_2
    \end{split}
\label{eq: lossbpr}
\end{equation}
where \textit{L} is the number of all ICD codes, $d_e$ represents the dimension of ICD codes, $n_j$ is the \textit{j-th} node of ICD codes, \textit{x} represents \textit{x-th} item of $n_j$, $\mathcal{L}_{bpr1}$ represents child-to-parent KL value between $n_j^{i_\uparrow}$ and $n_j^{i_\downarrow}$, $\mathcal{L}_{bpr2}$ represents parent-to-child KL value between $n_j^{i_\uparrow}$ and $n_j^{i_\downarrow}$.

When the loss value is miner than 0.01, the training stops. Finally, we obtain parent-child code representations \textbf{\emph{Vt}} $\in$ $\mathbb{R}^{d_e \times N}$.

\subsubsection{\textbf{Tree Position Encoder} (\textbf{\textit{TPE}})} \label{sec: spe}
Inspired by \cite{shiv2019novel}, we propose a tree position encoder to capture the positional features of ICD codes. 

\small
\begin{equation}
    PE_{\beta} = A_{\phi}PE_{\alpha}
\label{eq: treePEoverview}
\end{equation}

Given a position $\alpha$, we can get the target position $\beta$ with affine transform operation $A_{\phi}$ (Eq.(\ref{eq: treePEoverview})). The path between $\alpha$ and $\beta$ can be considered as the set of \emph{n} length-1 paths. Directions of these paths contains ``to parent" and ``to children". We take ``to parent" as function \emph{U} and ``to children" function \emph{D}. Thus, for any path $\phi$, we can obtain the transform $A_{\phi}$ by some combinations of \emph{U} and \emph{D}. For example, the position encoding of the second child of node \emph{x}'s grandpa can described path $\phi$ = $\langle$parent, parent, child-2$\rangle$, which can also be represented as $D_2U^2PE_x$.

We take the root node as zero vector (\textbf{0} $\in$ $\mathbb{R}^{d_e}$). Then, since all the paths from the root to other positions are downward, we can get embedding of any position \emph{x} via the Eq.(\ref{eq : tree-position}). 
\small
\begin{equation}
    \emph{x} = D_{b_L}D_{b_{L-1}}...D_{b_1}
    \label{eq : tree-position}
\end{equation}
where \emph{L} here means the \emph{L-th} layer the node is located, and $b_i$ represents the chosen path in the \emph{i-th} layer. We treat tree position encoding as a stack of length-1 component parts. Every \emph{D} operation pushes a length-1 path onto the stack, while \emph{U} pops a length-1 path, which is described as Eq.(\ref{eq: D_U}):

\small
\begin{equation}
    \begin{split}
        D_ix &= e^n_i \oplus (x\ominus x[n+1:]) \\
        Ux &= (x \ominus x[:(n-1)]) \oplus \textbf{0}_n
    \end{split}
    \label{eq: D_U}
\end{equation}
where $\ominus$ means pop operation or truncation, and $\oplus$ indicates push or concatenate operation. In addition, $e^n_i$ is an one-hot encoding with \emph{n} elements and \emph{n} is the total number of children of a parent. An example of function \textit{D} is shown in \ref{fig: treeposition}.

\piccaptioninside
\piccaption{An example of function \emph{D}}
\label{fig: treeposition}
\parpic[r]{\includegraphics[width=0.5\linewidth,keepaspectratio]{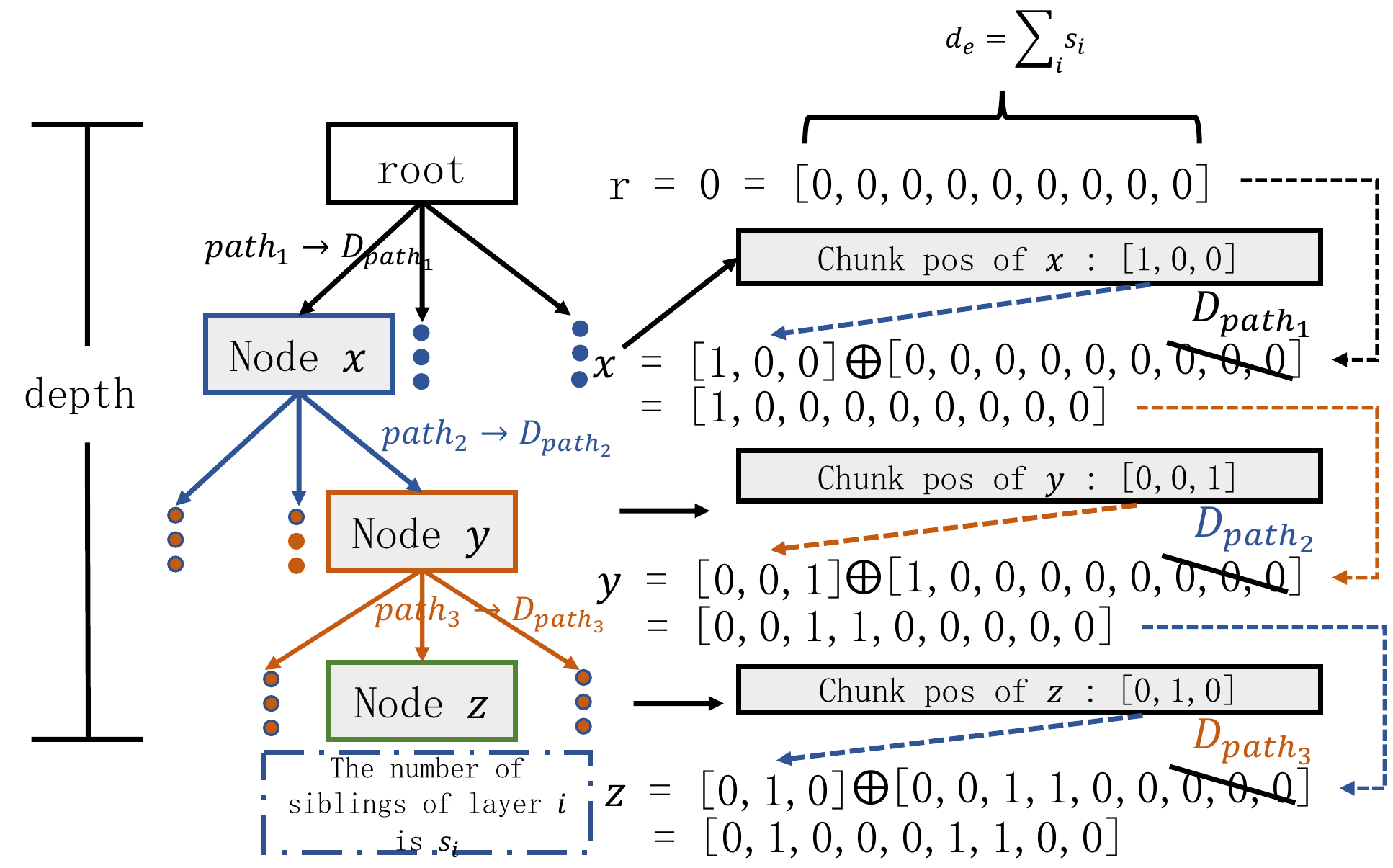}}

After tree-position encoding, we obtain code tree-position embedding (\textbf{$Vp$} $\in$ $\mathbb{R}^{d_e*N}$). Then add \textbf{\emph{Vp}} to \textbf{\emph{Vt}} to get final code representation $V_{pt}$ that contains both parent-child relationships and hierarchical features.

\subsubsection{\textbf{Code-Wise Attention}} \label{sec: codewise}
Inspired by \cite{cao-etal-2020-hypercore}, we use code-wise attention to generate code-aware document representations by using representations outputted from document feature extractor and bidirectional hierarchy passage encoder. The code-wise attention feature $a_l$ for code \textit{l} is calculated by:
\small
\begin{equation}
    \begin{split}
        s_l &= Softmax(\tanh(H\cdot W^T_a + b_a)\cdot {v_{pt}}_l) \\
        a_l &= s^T_l \cdot H \\
    \end{split}
\label{eq: code-wise}
\end{equation}
where \textit{Softmax} is the normalized exponential function, $s_l$ donates the attention scores for all elements in document representation \textit{H}, $a_l$ represents the most relevant information in \textit{H} about the code \textit{l} by code-wise attention. Then we get $d_e \times L$ dimensional code-wise adjacent document representations.

\subsection{Progressive Mechanism (\textit{PM}) for \textbf{\textit{DP}}} \label{sec: PM}

This section introduces a simple progressive mechanism to address the \textbf{\textit{DP}} problem in ICD coding. A former predicted code could help to predict next codes. For example, the neighbours of `Diabetes mellitus' contain `heart disease', neighbours of `heart disease contain `cardiovascular disease', then there exists a prediction path <Diabetes mellitus, heart disease, cardiovascular disease>. Given a clinical note, if `Diabetes mellitus' is true, then we could use this label to predict `heart disease'. Similarly, we can use `heart disease' to predict `cardiovascular disease'. But if `heart disease' is wrong, the process of progressive prediction will end. The problem is how to use the prior predicted code to predict other codes. We address the problem above by constructing an average-operation between the prior code's hidden output and other codes' hidden outputs with a hyper-parameter $\lambda$. Assuming the hidden outputs are described as $f_{i}\in \mathbb{R}^{1\times1}$ and it is proved true, then later hidden output $f_{j}$ should be calculated as follows:
\small
\begin{equation}
    \begin{split}
        f_j = \lambda f_i + (1-\lambda)*f_j
    \end{split}
\label{eq: progressive-eq}
\end{equation}
where $\lambda$ is a trade-off factor to balance $f_j$ and $f_i$. Here, $f_i$ decides the influence from a former code to a later code. $f_j$ reflects the current value of node \textit{j} itself. 

Figure \ref{fig: progressive_al} shows the dynamic progressive prediction in detail. After \emph{PM}, we obtain the output hidden embedding of ICD coding methods \emph{P} = \{$p_1$, $p_2$,..., $p_L$\} $\in$ $\mathbb{R}^{d_e*L}$. 

\subsection{Personalized PageRank (\textit{PP}) for \textbf{\textit{CC}}} \label{sec: PP}

Inspired by \emph{PPNP} \cite{klicpera2018predict}, we build a personalized PageRank \cite{page1999pagerank} to capture the code co-occurrence. Given a clinical text and its golden labels, we can build a strong connected sub-graph, and connect all sub-graphs into a big graph \emph{G = (V, E)}, where \emph{V} and \emph{E} are sets of nodes (codes) and edges respectively. Let \emph{L} denote the number of nodes and m the number of edges. The graph \emph{G} is described by the matrix \emph{A} $\in$ $\mathbb{R}^{L*L}$, and $\widetilde{A}$ = \emph{A} + $I_n$ denotes the matrix \emph{A} with added self-loops.

\piccaptioninside
\piccaption{Overview of progressive mechanism, \textit{Net} represents the network of method for automated ICD coding, \textit{BCE} is binary cross-entropy, \textit{GD} is short for Ground Truth.}
\label{fig: progressive_al}
\parpic[r]{\includegraphics[width=0.55\linewidth,keepaspectratio]{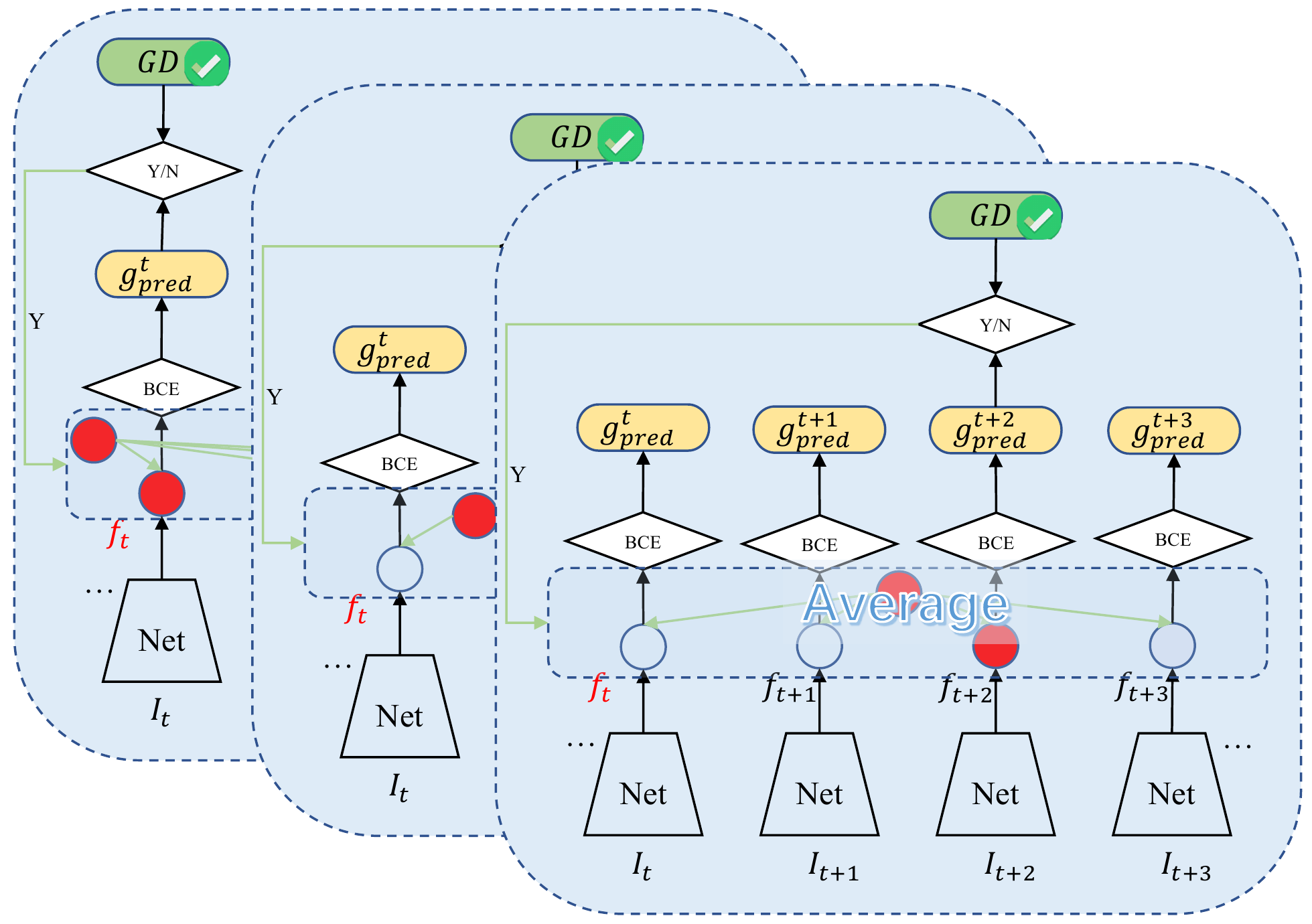}}

In the graph G, let the initial representation of node \emph{x} as $i_x$. In addition, We can update node embedding by the recursion equation:

\small
\begin{equation}
    \begin{aligned}
        PPR(i_x) &= (1-d) \hat{\widetilde{A}} + di_x \\ 
        & = d(I_n-(1-d)\hat{\widetilde{A}})^{-1}i_x
    \end{aligned}
    \label{eq: ppr}
\end{equation}
where $\hat{\widetilde{A}}$ = $\widetilde{D}^{-\frac{1}{2}}$ $\widetilde{A}$ $\widetilde{D}^{-\frac{1}{2}}$. \emph{D} is a degree matrix of nodes and $\widetilde{D}$ denotes the degree matrix of graph with added self-loops, where $\widetilde{D}_{i,i}$ = $\Sigma_j\widetilde{A}_{i,j}$

We generate predictions based on each node's own characteristics, and then propagate them through a fully personalized PageRank scheme to generate the pagerank-aware prediction. The pagerank-aware embedding is denoted as \emph{PPR} = \{$ppr_1$, $ppr_2$,..., $ppr_L$\} $\in$ $\mathbb{R}^{d_e*L}$.

\subsection{Aggregation and Training} \label{sec: aggregation}
\paragraph{Aggregation:}
After exploiting the structural information of code via \textbf{\textit{BHPE}} (i.e., \textbf{\textit{BPR}} and \textbf{\textit{TPE}}) and code co-occurrence via personalized PageRank, we obtain a normal representation (\emph{P}) (from \textbf{\textit{PM}} in Sec \ref{sec: PM}) and a pagerank-aware representation (\emph{PPR}) (Sec \ref{sec: PP}), respectively. We concatenate \emph{P} and \emph{PPR} as \emph{P3R} $\in$ $\mathbb{R}^{d_e*2L}$ and conduct a fully connected layer to reshape the matrix in to $\mathbb{R}^{d_e*L}$ embedding space. 

\paragraph{Training:}
Since the automated icd coding is a multi-label prediction task, we employ a multi-label binary cross-entropy (\textit{BCE}) as the loss of our model:
\small
\begin{equation}
    \begin{split}
        \mathcal{L}_{BCE}(y, \hat{y}) &= -\Sigma_{l=1}^L[y_l
        \log(\hat{y}_l) + (1-y_l)\log(1-\hat{y}_l)] 
    \end{split}
    \label{eq: loss}
\end{equation}
where $y_l$ is the ground truth and $\hat{y}_l$ is the predicted value, $\hat{y}_l = \sigma(x_l)$. Here, we use Adam optimizer\cite{DBLP:journals/corr/KingmaB14} to propagate the parameters of our model.

\section{Experimental Setup} \label{sec: exp}
\subsection{Datasets}

MIMIC-\uppercase\expandafter{\romannumeral2} \cite{jouhet2012automated} and MIMIC-\uppercase\expandafter{\romannumeral3} \cite{johnson2016mimic} are the most widely open-access datasets for evaluating automated ICD encoding methods. In MIMIC-\uppercase\expandafter{\romannumeral3}, there exists two versions. One is MIMIC-\uppercase\expandafter{\romannumeral3} full and the other is MIMIC-\uppercase\expandafter{\romannumeral3} 50. For MIMIC-\uppercase\expandafter{\romannumeral3} full, there are 8,921 unique codes, 47,723 discharge summaries for training, 3,372 summaries for test, and 1,631 for validation. For MIMIC-\uppercase\expandafter{\romannumeral3} 50, we use a set of 8,066 for training, with 1,729 summaries and 1,729 summaries for validation and test, respectively. In MIMIC-\uppercase\expandafter{\romannumeral2} dataset, there are 5,031 clinical codes, and we use the same experimental setting as previous works \cite{cao-etal-2020-hypercore, DBLP:journals/jamia/PerottePNWWE14, DBLP:journals/corr/abs-1709-09587}.

\subsection{Metrics and Parameter Settings}

We use macro/micro-averaged F1, macro/micro-averaged AUC, and P@N as the main metrics to evaluate our model and baselines. To measure \textbf{\textit{DP}}, we use Jaccard Similarity Coefficient \cite{niwattanakul2013using} as the metric, which is defined as $Jaccard = \frac{1}{m}\Sigma_i^m \lvert Y_i \cap \hat{Y_i}\rvert / \lvert Y_i \cup \hat{Y_i}\rvert$, where \textit{m} is the number of instances of the dataset, $Y_i$ is the predicted results from different ICD coding methods, $\hat{Y_i}$ indicates the ground truth ICD set (\textbf{note:} Assuming that len(Y) = $l_y$, len($\hat{Y}$ = \textit{K}, len(Y$\cap \hat{Y}$) = $l_{\cap}$, 
if \textit{K} $\ge$ $l_y$, let $\hat{Y}$ = $\hat{Y}[:m]$, else let Y = (Y $\cap \hat{Y}$) $\cup$ \textbf{0}*(\textit{K-$l_{\cap}$}).

We set the word embedding size $d_e$ as 100. The size of hidden layer is 128. We set 5 channels in CNN and the filer-sizes of them are 1, 3, 5, 7, 10, respectively. The dropout rate is 0.2. The learning rate is $1e^{-4}$. The batch size is set as 32. The dump rate \emph{d} in Eq.(\ref{eq: ppr}) is 5. The maximize of personalized PageRank loop is 50. The criterion for early stopping is 10. The initial embedding tool for clinical codes is word2vec \cite{DBLP:journals/corr/abs-1301-3781}. The initial embedding tool for clinical text is WordPiece tokenizer in BERT. The reason of why we use different initial embedding methods for clinical codes and nodes are shown in footnote \textit{1} in Sec\ref{sec: cnn}. 

\subsection{Baselines}
In order to demonstrate the effectiveness of HieNet, we compare it with several previous methods, including state-of-the-art models with knowledge graph and GCN.
\begin{itemize}
    \item [$\bullet$] \textbf{CNN}. CNN is a widely applied method in language modeling \cite{DBLP:conf/icml/DauphinFAG17}.
    \item [$\bullet$] \textbf{CAML \& DRCAML} \cite{mullenbach-etal-2018-explainable}. CAML leverages convolutional attention for automated ICD prediction. DR-CAML is a updated version of CAML with code description.
    \item [$\bullet$] \textbf{HyperCore} \cite{cao-etal-2020-hypercore}. This method employed hyperbolic representation to capture the code hierarchy and used GCN \cite{DBLP:conf/iclr/KipfW17} to encode the semantic of the code co-occurrence.
    \item [$\bullet$] \textbf{JointLATT} \cite{DBLP:journals/corr/abs-2007-06351}. This method proposed a new label attention method. And the label attention model achieve SOTA results compared with other previous works.
    \item [$\bullet$] \textbf{MASTATT-KG} \cite{xie2019ehr}. This method utilizes a multi-scale feature attention to select multi-scale features adaptively. 
    \item [$\bullet$] \textbf{MultiResCNN} \cite{li2020icd}. MultiResCNN contains a multi-filter convolutional layer to capture various text patterns and a residual convolutional layer to enlarge the receptive field.
    \item [$\bullet$] \textbf{DCAN}\cite{ji2020dilated}. This method proposes a dilated convolutional attention network, integrating dilated convolutions, residual connections, and label attention, for medical code assignment.
\end{itemize}

\section{Result and Analysis}
We focus on answering the following \textbf{researching questions} (\textbf{RQs}):

\begin{itemize}
    \item [$\bullet$] \textbf{RQ1:} What is the performance of HieNet on ICD encoding task? 
    \item [$\bullet$] \textbf{RQ2:} Why progressive mechanism and how it performs?
    \item [$\bullet$] \textbf{RQ3:} What are the contributions of the different components?
    \item [$\bullet$] \textbf{RQ4:} How does the trade-off coefficient ($\lambda$ in Sec \ref{sec: PM}) influence the performance?
\end{itemize}
\subsection{Overall Performance (RQ1)}

\begin{table}[]
\centering
\caption{Results (\%) of the comparison of our model and other baselines on the MIMIC-\uppercase\expandafter{\romannumeral3} full and MIMIC-\uppercase\expandafter{\romannumeral3} 50. In all tables, the bold number with * indicates the best result compared to other methods.}
\resizebox{0.9\textwidth}{!}{
\begin{tabular}{|l|cccccccc|}
\hline
 &
  \multicolumn{8}{c|}{MIMIC-III full} \\ \hline
\multirow{2}{*}{Model} &
  \multicolumn{2}{c|}{Jaccard} &
  \multicolumn{2}{c|}{AUC} &
  \multicolumn{2}{c|}{F1} &
  \multicolumn{2}{c|}{P@N} \\ \cline{2-9} 
 &
  \multicolumn{1}{c|}{top\_20} &
  \multicolumn{1}{c|}{top\_30} &
  \multicolumn{1}{c|}{Macro} &
  \multicolumn{1}{c|}{Micro} &
  \multicolumn{1}{c|}{Macro} &
  \multicolumn{1}{c|}{Micro} &
  \multicolumn{1}{c|}{8} &
  15 \\ \hline
CNN &
  \multicolumn{1}{c|}{30.2} &
  \multicolumn{1}{c|}{20.9} &
  \multicolumn{1}{c|}{80.6} &
  \multicolumn{1}{c|}{96.9} &
  \multicolumn{1}{c|}{4.2} &
  \multicolumn{1}{c|}{41.9} &
  \multicolumn{1}{c|}{40.2} &
  49.1 \\ \hline
CAML &
  \multicolumn{1}{c|}{32.4} &
  \multicolumn{1}{c|}{22.5} &
  \multicolumn{1}{c|}{88.8} &
  \multicolumn{1}{c|}{98.4} &
  \multicolumn{1}{c|}{7.1} &
  \multicolumn{1}{c|}{51.9} &
  \multicolumn{1}{c|}{69.7} &
  54.9 \\ \hline
DR-CAML &
  \multicolumn{1}{c|}{33.7} &
  \multicolumn{1}{c|}{23.1} &
  \multicolumn{1}{c|}{89.7} &
  \multicolumn{1}{c|}{98.5} &
  \multicolumn{1}{c|}{8.6} &
  \multicolumn{1}{c|}{52.9} &
  \multicolumn{1}{c|}{69.0} &
  54.8 \\ \hline
HyperCore &
  \multicolumn{1}{c|}{-} &
  \multicolumn{1}{c|}{-} &
  \multicolumn{1}{c|}{93.0} &
  \multicolumn{1}{c|}{98.9} &
  \multicolumn{1}{c|}{9.0} &
  \multicolumn{1}{c|}{55.1} &
  \multicolumn{1}{c|}{72.2} &
  57.9 \\ \hline
JointLAAT &
  \multicolumn{1}{c|}{-} &
  \multicolumn{1}{c|}{-} &
  \multicolumn{1}{c|}{92.1} &
  \multicolumn{1}{c|}{98.8} &
  \multicolumn{1}{c|}{8.9} &
  \multicolumn{1}{c|}{55.3} &
  \multicolumn{1}{c|}{73.5} &
  59.0 \\ \hline
MSATT-KG &
  \multicolumn{1}{c|}{32.1} &
  \multicolumn{1}{c|}{22.0} &
  \multicolumn{1}{c|}{91.0} &
  \multicolumn{1}{c|}{99.2} &
  \multicolumn{1}{c|}{9.0} &
  \multicolumn{1}{c|}{55.3} &
  \multicolumn{1}{c|}{72.8} &
  58.1 \\ \hline
MultiResCNN &
  \multicolumn{1}{c|}{-} &
  \multicolumn{1}{c|}{-} &
  \multicolumn{1}{c|}{91.0} &
  \multicolumn{1}{c|}{98.6} &
  \multicolumn{1}{c|}{8.5} &
  \multicolumn{1}{c|}{55.2} &
  \multicolumn{1}{c|}{73.4} &
  58.4 \\ \hline
HieNet &
  \multicolumn{1}{c|}{\textbf{36.3*} $\pm$ 0.2} &
  \multicolumn{1}{c|}{\textbf{27.5*} $\pm$ 0.3} &
  \multicolumn{1}{c|}{\textbf{93.3*} $\pm$ 0.4} &
  \multicolumn{1}{c|}{\textbf{99.2*} $\pm$ 0.2} &
  \multicolumn{1}{c|}{\textbf{9.3*} $\pm$ 0.1} &
  \multicolumn{1}{c|}{\textbf{56.6*} $\pm$ 0.7} &
  \multicolumn{1}{c|}{\textbf{78.3*} $\pm$ 0.5} &
  65.0* 0.3 \\ \hline\hline
 &
  \multicolumn{8}{c|}{MIMIC-III 50} \\ \hline
\multirow{2}{*}{Model} &
  \multicolumn{2}{c|}{Jaccard} &
  \multicolumn{2}{c|}{AUC} &
  \multicolumn{2}{c|}{F1} &
  \multicolumn{2}{c|}{P@N} \\ \cline{2-9} 
 &
  \multicolumn{1}{c|}{top\_20} &
  \multicolumn{1}{c|}{top\_30} &
  \multicolumn{1}{c|}{Macro} &
  \multicolumn{1}{c|}{Micro} &
  \multicolumn{1}{c|}{Macro} &
  \multicolumn{1}{c|}{Micro} &
  \multicolumn{1}{c|}{5} &
  - \\ \hline
CNN &
  \multicolumn{1}{c|}{31.7} &
  \multicolumn{1}{c|}{24.3} &
  \multicolumn{1}{c|}{87.6} &
  \multicolumn{1}{c|}{90.7} &
  \multicolumn{1}{c|}{57.6} &
  \multicolumn{1}{c|}{62.5} &
  \multicolumn{1}{c|}{62.0} &
  - \\ \hline
CAML &
  \multicolumn{1}{c|}{32.5} &
  \multicolumn{1}{c|}{25.2} &
  \multicolumn{1}{c|}{87.5} &
  \multicolumn{1}{c|}{90.9} &
  \multicolumn{1}{c|}{53.2} &
  \multicolumn{1}{c|}{61.4} &
  \multicolumn{1}{c|}{60.9} &
  - \\ \hline
DR-CAML &
  \multicolumn{1}{c|}{32.5} &
  \multicolumn{1}{c|}{24.6} &
  \multicolumn{1}{c|}{88.4} &
  \multicolumn{1}{c|}{91.6} &
  \multicolumn{1}{c|}{57.6} &
  \multicolumn{1}{c|}{63.3} &
  \multicolumn{1}{c|}{61.8} &
  - \\ \hline
HyperCore &
  \multicolumn{1}{c|}{-} &
  \multicolumn{1}{c|}{-} &
  \multicolumn{1}{c|}{89.5} &
  \multicolumn{1}{c|}{92.9} &
  \multicolumn{1}{c|}{60.9} &
  \multicolumn{1}{c|}{66.3} &
  \multicolumn{1}{c|}{63.2} &
  - \\ \hline
JointLAAT &
  \multicolumn{1}{c|}{-} &
  \multicolumn{1}{c|}{-} &
  \multicolumn{1}{c|}{92.5} &
  \multicolumn{1}{c|}{94.6} &
  \multicolumn{1}{c|}{66.1} &
  \multicolumn{1}{c|}{71.6} &
  \multicolumn{1}{c|}{67.1} &
  - \\ \hline
MSATT-KG &
  \multicolumn{1}{c|}{33.9} &
  \multicolumn{1}{c|}{24.7} &
  \multicolumn{1}{c|}{91.4} &
  \multicolumn{1}{c|}{93.6} &
  \multicolumn{1}{c|}{63.8} &
  \multicolumn{1}{c|}{68.4} &
  \multicolumn{1}{c|}{64.4} &
  - \\ \hline
MultiResCNN &
  \multicolumn{1}{c|}{-} &
  \multicolumn{1}{c|}{-} &
  \multicolumn{1}{c|}{89.9} &
  \multicolumn{1}{c|}{92.8} &
  \multicolumn{1}{c|}{60.6} &
  \multicolumn{1}{c|}{67.0} &
  \multicolumn{1}{c|}{64.1} &
  - \\ \hline
DCAN &
  \multicolumn{1}{c|}{-} &
  \multicolumn{1}{c|}{-} &
  \multicolumn{1}{c|}{90.2} &
  \multicolumn{1}{c|}{93.1} &
  \multicolumn{1}{c|}{61.5} &
  \multicolumn{1}{c|}{67.1} &
  \multicolumn{1}{c|}{64.2} &
  - \\ \hline
HieNet &
  \multicolumn{1}{c|}{\textbf{37.7*} $\pm$ 0.1} &
  \multicolumn{1}{c|}{\textbf{30.4*} $\pm$ 0.2} &
  \multicolumn{1}{c|}{\textbf{93.4*} $\pm$ 0.8} &
  \multicolumn{1}{c|}{\textbf{95.0*} $\pm$ 0.1} &
  \multicolumn{1}{c|}{\textbf{67.1*} $\pm$ 0.2} &
  \multicolumn{1}{c|}{\textbf{72.4*} $\pm$ 0.2} &
  \multicolumn{1}{c|}{\textbf{69.5*} $\pm$ 0.3} &
  - \\ \hline
\end{tabular}}
\label{tb: mimic3}
\end{table}

The comparisons between our model and other state-of-the-art on MIMIC-\uppercase\expandafter{\romannumeral2} and MIMIC-\uppercase\expandafter{\romannumeral3} are given in Table \ref{tb: mimic3} and Table \ref{tb: mimic2}, respectively. Our HieNet model outperforms every single baseline on most of metrics. The CAML architecture is comparable to the DR-CAML, and the CNN baseline essentially performs the worst than all other neural architectures. We recognized P@N as the most intuitive measure to indicate the effectiveness of methods, since it examines the ability of the method to return a high-confidence subset of codes. Moreover, Jaccard is used to measure the accuracy of $top_K$ predicted codes.

\textbf{For MIMIC-\uppercase\expandafter{\romannumeral3} full:}
Compared with baselines, HieNet achieves the best perfermance on macro-F1, micro-F1, and macro-AUC. Since clinical codes are in uneven distribution and macro-F1 emphasizes the performance of rare label, it is difficult to obtain high macro-F1 score. Even in this case, HieNet performs perfectly and achieves 3\% improvement compared to the latest state-of-the-art HyperCore method. This demonstrates the effectiveness of HieNet. Furthermore, on Jaccard metric, HieNet improves the performance by a big margin with 7.7\% improvement on $top_{20}$ (from 33.7\% to 36.3\%) and 19.0\% improvement on $top_{30}$ (from 23.1\% to 27.5\%).

\textbf{For MIMIC-\uppercase\expandafter{\romannumeral3} 50:}
Following the previous work \cite{mullenbach-etal-2018-explainable, cao-etal-2020-hypercore}, we also evaluate our model and baselines on the most common 50 codes set of MIMIC-\uppercase\expandafter{\romannumeral3}. Different from MIMIC-\uppercase\expandafter{\romannumeral3} full, MIMIC-\uppercase\expandafter{\romannumeral3} 50 has a relatively smooth distribution, which leads to the possibility of achieving higher macro-F1 scores.

Our method obtains the highest score on macro-AUC, micro-AUC, and P@5 metrics. On MIMIC-\uppercase\expandafter{\romannumeral3} 50, HieNet achieves the best performance on most of the evaluation except micro-F1 and macro-F1. The $top_{30}$-Jaccard value gets the most significant improvement (i.e., 20.6\% over the best baseline CAML). The reason is that codes on datasets on MIMIC-\uppercase\expandafter{\romannumeral3} 50 are more closely related, and progressive mechanism (\textit{CC}) is just designed for this.

\begin{table}[]
\centering
\caption{Experimental results of our model and other baselines on MIMIC-\uppercase\expandafter{\romannumeral2}.}
\resizebox{0.8\textwidth}{!}{
\begin{tabular}{|l|c|c|c|c|c|c|c|}
\hline
\multirow{2}{*}{\textbf{Model}} & \multicolumn{2}{c|}{\textbf{Jaccard}} & \multicolumn{2}{c|}{\textbf{AUC}} & \multicolumn{2}{c|}{\textbf{F1}} & \textbf{P@N} \\ \cline{2-8} 
            & $top_{20}$   & $top_{30}$   & Macro & Micro & Macro & Micro & 8    \\ \hline
CNN         & 13.7 & 11.3 & 74.2  & 94.1  & 3.0   & 33.2  & 38.8 \\ \hline
CAML        & 14.4 & 12.3 & 82.0  & 96.6  & 4.8   & 44.2  & 52.3 \\ \hline
DR-CAML     & 13.8 & 11.7 & 82.6  & 96.6  & 4.9   & 45.7  & 51.1 \\ \hline
HyperCore   & -    & -    & 88.5  & 97.1  & 7.0   & 47.7  & 53.7 \\ \hline
JointLAAT   & -    & -    & 87.1  & 97.2  & 6.8   & 49.1  & 55.1 \\ \hline
MultiResCNN & -    & -    & 85.1  & 96.8  & 5.2   & 46.4  & 54.4 \\ \hline
HieNet                          & \textbf{15.7*} $\pm$ 0.1          & \textbf{13.2*} $\pm$ 0.3          & \textbf{89.4*} $\pm$ 0.2        & \textbf{98.3*} $\pm$ 0.1        & \textbf{7.1*} $\pm$ 0.5        & \textbf{49.2*} $\pm$ 0.2        & \textbf{56.6*} $\pm$ 0.3     \\ \hline
\end{tabular}}
\label{tb: mimic2}
\end{table}

\textbf{For MIMIC-\uppercase\expandafter{\romannumeral2}}:
As shown in Table \ref{tb: mimic2}, MIMIC-\uppercase\expandafter{\romannumeral2} contains 5,031 labels, and our method HieNet also performs the best on most metrics compared with baselines except macro-F1 value. In addition, latest work HyperCore's macro-AUC, micro-AUC, micro-F1, and P@8 are much lower than HieNet (1.0\%, 1.2\%, 3.1\%, and 5.4\% lower). $top_{20}$ and $top_{30}$ Jaccard values indicate that CNN, CAML, and DR-CAML are poor in leveraging relationships among codes while HieNet can predicts a higher coverage of correct ICD codes.

\subsection{Effectiveness of Progressive Mechanism (RQ2)}

\begin{figure}[]
\centerline{\includegraphics[width=0.9\linewidth,keepaspectratio]{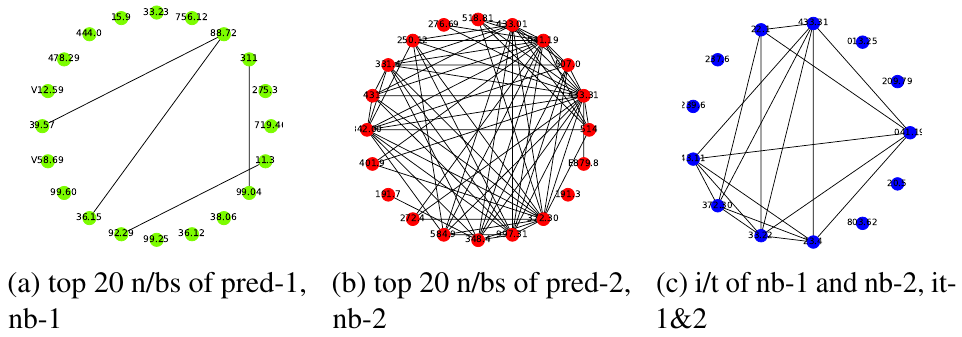}}
\caption{Example of progressive mechanism in HieNet on patient 118299 summaries. n/bs means neighbors, i/t represents interaction, pred-1 indicates the top 1 of predicted codes, nb-\textit{i} denotes the set of neighbors of \textit{i-th} predicted codes, and it-1\&2 represents the interacted result of nb-1 and nb-2.}
\label{fig: progressive}
\end{figure}

We take an example in Figure \ref{fig: progressive}. For a given clinical text (patient 118299 here), we verify the impact of the first predicted code on predicting the second one, and the effect of the front two predicted codes on predicting the third one.

The gold label of patient 118299 is \emph{gl} = [`198.3', `348.5', `162.9', `401.9', `272.4', `15.9', `22.0', `20.5'], and the front predicted codes are `401.9' and `15.9'. Node `401.9' has 1,655 neighbours and Figure 5a just shows the 20 of them. Node `15.9' has 56 neighbors and Figure 5b only shows 20 of them. The shared neighbors (\emph{i/t-1\&2} in Figure 5c) are \emph{i/t-1\&2} = [`041.19', `22.1', `433.31', `209.79', `013.25', `237.6', `239.6', `43.11', `372.30', `33.22', `23.4', `803.62', `20.5']. As observed, `20.5' occurred in \emph{i/t-1\&2} is exactly one element in \emph{gl}. Thus, we only need 13 labels (\emph{i/t-1\&2}) to predict the third code.

Obviously, the neighbors of the former predicted codes can be used help predict the one. In conclusion, the process of above demonstrates the effectiveness of progressive mechanism.

\subsection{Ablation Study (RQ3)}
We conduct ablation investigation to examine the effectiveness of each part in our model. To evaluate a model, we remove it (denoted as without, w/o) and perform the remaining part on the datasets. The experimental results of ablation study are shown in Table \ref{tb: ablation}.

\begin{table*}[ht]
\centering
\caption{Ablation study by removing the main components.}
\resizebox{0.8\textwidth}{!}{
\begin{tabular}{lcccccc}
\hline
\multirow{2}{*}{Models}           & \multicolumn{2}{c}{MIMIC-\uppercase\expandafter{\romannumeral3} full} & \multicolumn{2}{c}{MIMIC-\uppercase\expandafter{\romannumeral3} 50} & \multicolumn{2}{c}{MIMIC-\uppercase\expandafter{\romannumeral2}}   \\ \cline{2-7}  
 &
  \multicolumn{1}{l}{Macro-F1} &
  \multicolumn{1}{l}{Micro-F1} &
  \multicolumn{1}{l}{Macro-F1} &
  \multicolumn{1}{l}{Micro-F1} &
  \multicolumn{1}{l}{Macro-F1} &
  \multicolumn{1}{l}{Micro-F1} \\ \hline 
HieNet                             & \textbf{9.3*}   & \textbf{56.6*}   & \textbf{67.1*}  & \textbf{72.4*}  & \textbf{6.9*} & \textbf{49.2*} \\ 
w/o \textit{PM}           & 8.9            & 56.3            & 59.8           & 64.1           & 6.5          & 47.7          \\  
w/o \textit{BHPE}  & 8.3            & 55.2            & 57.7           & 62.6           & 6.1          & 47.5          \\ 
w/o \textit{PP}  & 8.4            & 56.1            & 58.6          & 65.3           & 6.5          & 47.3       \\ 
w GCN for \textit{CC} (w/o \textit{PP})  & 9.1            & 55.3            & 60.4          & 65.7           & 6.7          & 49.0      \\
w tree-lstm for \textit{DP} (w/o \textit{BHPE} )  & 8.8            & 55.2            & 60.0         & 64.6           & 6.6          & 47.9      \\
\hline
\end{tabular}}
\label{tb: ablation}
\end{table*}

\textbf{Impact of} \textbf{\textit{PM}}. We remove the \textit{PM} part from the full model. As shown in Table \ref{tb: ablation}, HieNet without progressive module achieves lower scores of macro-F1, micro-F1 on both MIMIC-\uppercase\expandafter{\romannumeral2} and MIMIC-\uppercase\expandafter{\romannumeral3}.

\textbf{Impact of} \textbf{\textit{PP}}. Compared with the w/o personalized PageRank module, the full HieNet improves the score on Macro-F1 (MIMIC-\uppercase\expandafter{\romannumeral3} full) from 0.084 to 0.093, 0.586 to 0.611 (MIMIC-\uppercase\expandafter{\romannumeral3} 50), and 0.065 to 0.069 (MIMIC-\uppercase\expandafter{\romannumeral2}), respectively. As shown in Table \ref{tb: ablation}, the line `w GCN for \textit{CC} (w/o \textit{PP})' performs poorer than HieNet.

\textbf{Impact of} \textbf{\textit{BHPE}}. We remove \textit{BHPE} and compare with the full HieNet. The result is given in Table \ref{tb: ablation}, HieNet improves 10.7\% (from 0.084 to 0.093) on macro-F1 (MIMIC-\uppercase\expandafter{\romannumeral3} full), and 14.5\% (from 0.586 to 0.671) on macro-F1 (MIMIC-\uppercase\expandafter{\romannumeral3} 50), 6.2\% (from 0.065 to 0.069) on macro-F1 (MIMIC-\uppercase\expandafter{\romannumeral2}), etc, respectively. 

Table \ref{tb: ablation} shows that HieNet with tree-lstm instead of \textit{BHPE} performs worse on all metrics on three main datasets. The results from Table \ref{tb: ablation} demonstrates the effectiveness of different modules in HieNet. In addition, \textit{BHPE} module plays a more important role in HieNet compared with \textit{PM} and \textit{PP}.

\subsection{The impact of $\lambda$ (RQ4)}

\piccaptioninside
\piccaption{Performance with different values of $\lambda$.}
\label{fig: lamda}
\parpic[r]{\includegraphics[width=0.7\linewidth,keepaspectratio]{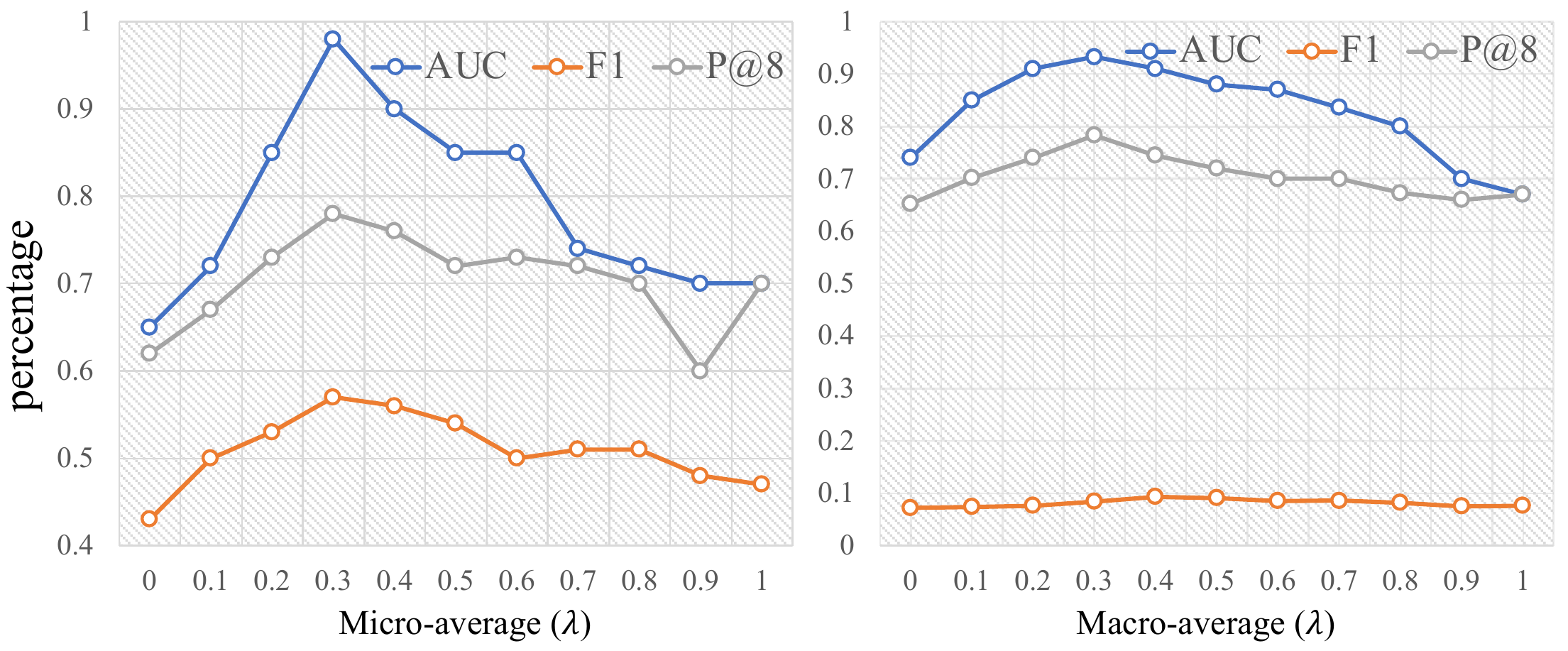}}
The trade-off parameter $\lambda$ is used to balance the influence factor from former hidden outputs and current hidden output. When $\lambda$ is larger, HieNet relies more on the former predicted codes to influence the next ones to predict, which means there exists a strong reasoning relationship in the predicted codes corresponding to the given clinical note. When $\lambda$ is smaller, HieNet tends to prioritize the current hidden output to predict a proper ICD code during the learning. The value of $\lambda$ is set from (0, 0.1, 0.2, ..., 1.0) to measure the performance. The results measured on MMIMIC-\uppercase\expandafter{\romannumeral3}-full are shown in Figure \ref{fig: lamda}.

First, as $\lambda$ increases, all metrics increase gradually at the beginning, but most of them decrease when $\lambda$ $\ge$ 0.3. In addition, F1 of macro-average decreases after $\lambda$ $\ge$ 0.4. The best performance is not achieved when $\lambda$ = 0 or 1. This demonstrates that progressive mechanism improve the performance of ICD coding methods. Second, Figure \ref{fig: lamda} shows that the beginning increases of macro-F1, macro-AUC, micro-F1, and micro-AUC are faster than the afterward decreases. Specially, micro-AUC grows from 0.65 to 0.98 only by the 0.3 added $\lambda$-value while micro-AUC drops from 0.98 to 0.7 needs another 0.7 added $\lambda$-value. This indicates the information of prior predicted codes can significantly affect the performance of ICD coding.

\section{Conclusion and Future Works}
In this paper, we propose the HieNet, which employs multi-channel CNN to encode the document representation, bidirectional hierarchy encoder to capture the hierarchy features, progressive mechanism improve the accuracy of $top_K$ predicted codes, and personalized PageRank to obtain code co-occurrence. HieNet yields strong improvements over previous methods, while providing the new state-of-the-art performance on both MIMIC-\uppercase\expandafter{\romannumeral2} and MIMIC-\uppercase\expandafter{\romannumeral3}.

However, for input, the future works need to pay more attention to multi-type input. The input can be medical image (i.e., chest radio graph), structured information (i.e., prescriptions), unstructured data (i.e., clinical texts), etc. Inspired by the effectiveness of the progressive mechanism, we can build a multi-model prediction model in future.

\bibliographystyle{splncs04}
\bibliography{sample-base.bib}
\end{document}